\documentclass[11pt,a4paper]{article}
\usepackage[hyperref]{emnlp2018}
\usepackage{times}
\usepackage{latexsym}
\usepackage{amsfonts}
\usepackage{bbm}
\usepackage{graphicx}
\usepackage{url}
\usepackage{amsmath}
\usepackage{todonotes}

\newcounter{dm}

\newcounter{mw}

\newcounter{sy}

\newcounter{rb}

\usepackage{caption}
\DeclareCaptionFormat{capfont}{\fontsize{10}{6}\selectfont#1#2#3}
\usepackage{subcaption}
\usepackage{algorithmicx} 
\usepackage[noend]{algpseudocode}
\usepackage{algorithm}
\usepackage[T1]{fontenc}

\algnewcommand\algorithmicdefinitions{\textbf{Definitions:}}
\algnewcommand\Definitions{\item[\algorithmicdefinitions]}
\renewcommand{\algorithmiccomment}[1]{{\color{gray}\raisebox{1px}{\texttt{\guillemotright}} #1}}
\algnewcommand{\LineComment}[1]{\Statex \hskip\ALG@thistlm \algorithmiccomment{#1}}
\algrenewcommand\alglinenumber[1]{\footnotesize #1:}
\algrenewcommand\algorithmicindent{1.0em}%
\makeatletter
\newcommand{\StatexIndent}[1][3]{%
  \setlength\@tempdima{\algorithmicindent}%
  \Statex\hskip\dimexpr#1\@tempdima\relax}

\aclfinalcopy %
\def\aclpaperid{2214} %

\newcommand{\nlstring}[1]{{\em #1}}

\newcommand{\policy}{p}

\newcommand{\programset}{\mathcal{Y}}
\newcommand{\env}{e}
\newcommand{\answer}{z}

\newcommand{\tab}{t}
\newcommand{\exec}{\Phi}
\newcommand{\predicate}{\omega}
\newcommand{\token}{w}
\newcommand{\score}{{\tt score}}
\newcommand{\program}{y}

\newcommand{\instruction}{x}
\newcommand{\allinstruction}{\mathcal{X}}
\newcommand{\alltable}{\mathcal{T}}

\newcommand{\allanswer}{\mathcal{Z}}

\DeclareMathOperator*{\argmax}{arg\,max}

\usepackage[printwatermark]{xwatermark}
\usepackage{tikz}

\title{Policy Shaping and Generalized Update Equations for \\ Semantic Parsing from Denotations}

\author{Dipendra Misra$^\star$, Ming-Wei Chang$^{\dagger}$, Xiaodong He$^{\diamond}$, Wen-tau Yih$^{\ddagger}$ \\
	$^\star$Cornell University, $^\dagger$Google AI Language, $^\diamond$JD AI Research\\
	$^\ddagger$Allen Institute for Artificial Intelligence\\
\texttt{dkm@cs.cornell.edu, mingweichang@google.com} \\
\texttt{xiaodong.he@jd.com, scottyih@allenai.org}
}

\date{}

\begin{document}
\maketitle

  \begin{abstract}

Semantic parsing from denotations faces two key challenges in model training: (1) given only the denotations (e.g., answers), search for good candidate semantic parses, and (2) choose the best model update algorithm.
We propose effective and general solutions to each of them.
Using \emph{policy shaping}, we bias the search procedure towards semantic parses that are more compatible to the text, which provide better supervision signals for training.
In addition, we propose an update equation that generalizes three different families of learning algorithms, which enables fast model exploration. 
When experimented on a recently proposed sequential question answering dataset, our framework leads to a new state-of-the-art model that outperforms previous work by 5.0\% absolute on exact match accuracy.

\end{abstract}

  \section{Introduction}  %

Semantic parsing from denotations (SpFD) is the problem of mapping text to executable formal representations (or {\em program}) in a situated environment and executing them to generate denotations (or {\em answer}), in the absence of access to correct representations. Several problems have been handled within this framework, including question answering~\cite{Berant:13,Iyyer:17seq-qa} and instructions for robots ~\cite{Artzi:13,Misra:15highlevel}. %

\begin{figure}
\centering
\includegraphics[scale=0.28]{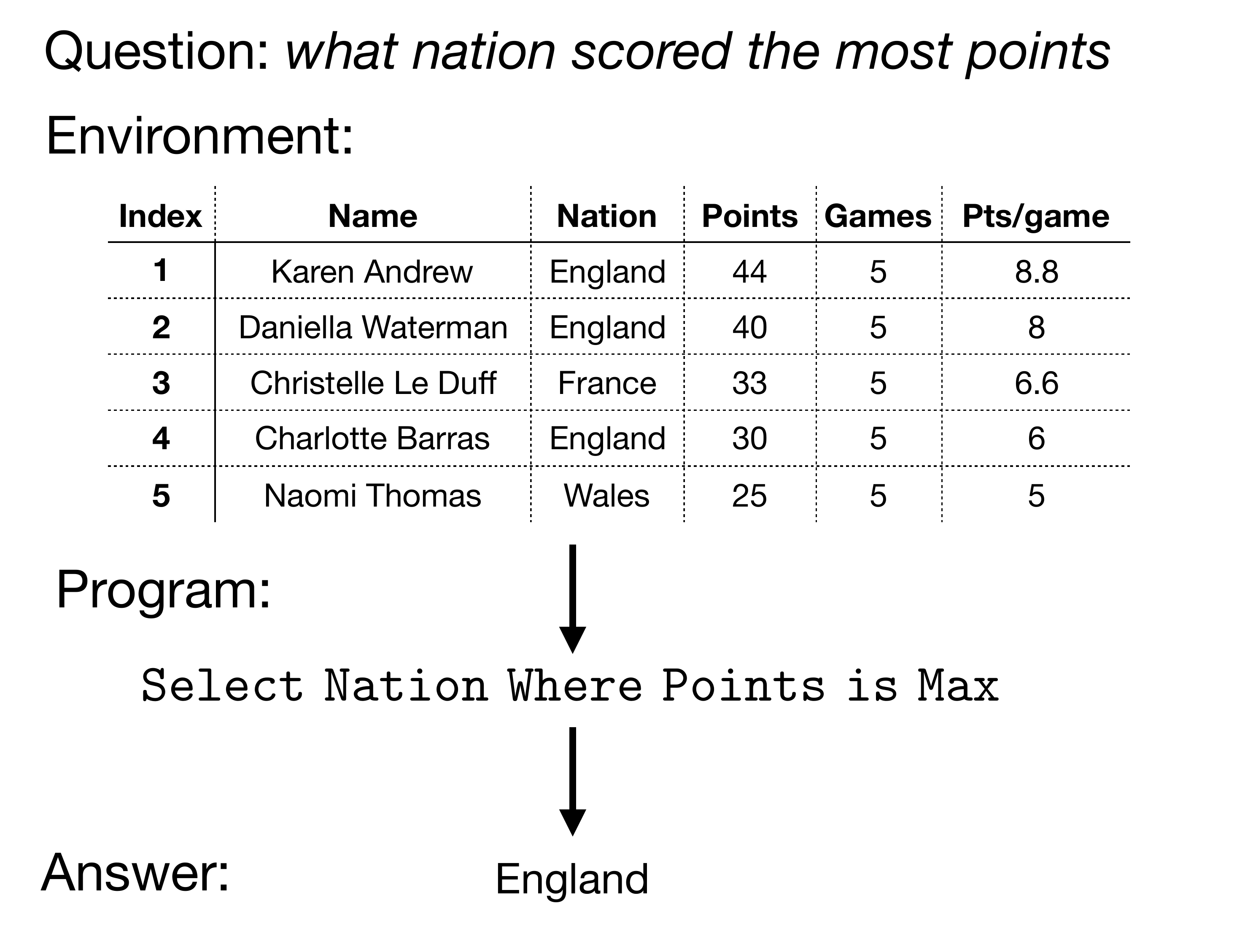}

\caption{An example of semantic parsing from denotations. Given the table environment, map the question to an executable program that evaluates to the answer.}
\label{fig:example}
\end{figure}

\begin{figure*}
\centering
\includegraphics[scale=0.2]{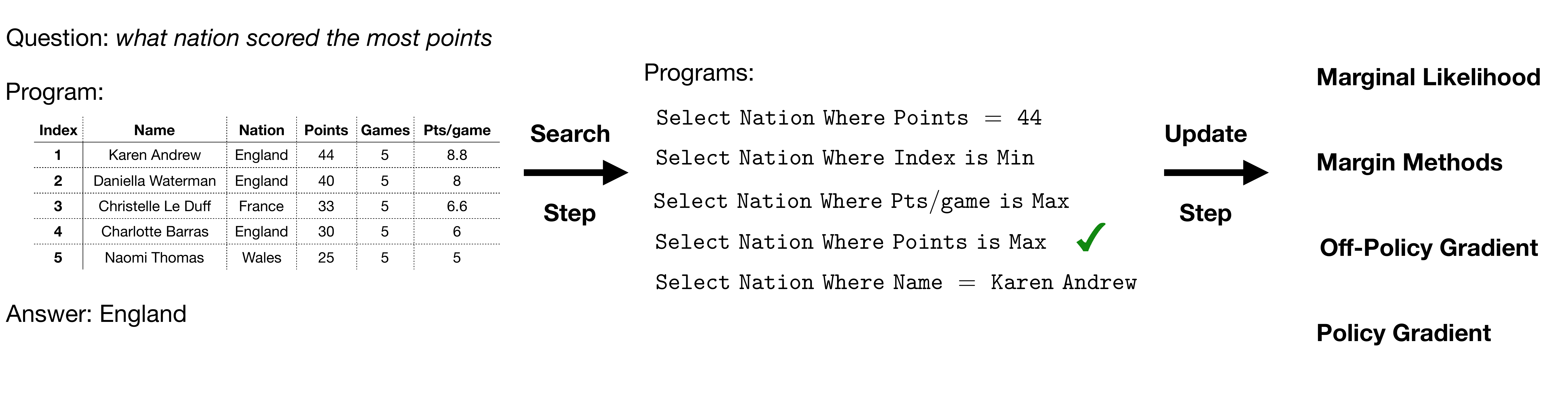}
\caption{An example of semantic parsing from denotation. Given the question and the table environment, there are several programs which are spurious.}
\label{fig:flowchart}
\end{figure*}

Consider the example in Figure~\ref{fig:example}. Given the question and a table environment, a semantic parser maps the question to an executable program, in this case a SQL query, and then executes the query on the environment to generate the answer \nlstring{England}. In the SpFD setting, the training data does not contain the correct programs. Thus, the existing learning approaches for SpFD perform two steps for every training example, a search step that explores the space of programs and finds suitable candidates, and an update step that uses these programs to update the model. Figure~\ref{fig:flowchart} shows the two step training procedure for the above example.

In this paper, we address two key challenges in model training 
for SpFD by proposing a novel learning framework, improving both
the \emph{search} and \emph{update} steps.
The first challenge, the existence of
{\em spurious programs}, lies in the search step. More specifically,
while the success of the search step relies on its ability to find programs that are semantically correct, we
can only verify if the program can generate correct
answers, given that no gold programs are presented. The search step is complicated by \emph{spurious programs}, which 
happen to evaluate to the correct answer but do not represent accurately the meaning of
the natural language question. For example, for the environment in Figure~\ref{fig:example}, the program {\tt Select Nation Where Name = Karen Andrew} is spurious. Selecting spurious programs as positive examples can greatly affect the performance of semantic parsers as these programs generally do not generalize to unseen questions and environments. %

The second challenge, {\em choosing a learning algorithm}, lies in the update step. Because of the unique \emph{indirect supervision} setting of SpFD, the quality
of the learned semantic parser is dictated by the choice of how to update the model parameters, often determined empirically. 
As a result, several families of learning methods, including maximum marginal likelihood, reinforcement learning and margin based methods have been used. How to effectively explore different model choices could be crucial in practice.

Our contributions in this work are twofold. To address the first challenge, we propose a policy shaping~\cite{Griffith:13policyshaping} method that incorporates simple, lightweight domain knowledge, such as a small set of lexical pairs of tokens in the question and program, in the form of a \emph{critique policy} (\S~\ref{sec:search}). This helps bias the search towards the correct program, an important step to improve supervision signals, which benefits learning regardless of the choice of algorithm. 
To address the second challenge, 
we prove that the parameter update step in several algorithms are similar and can be viewed as special cases of a generalized update equation (\S~\ref{sec:learning}). 
The equation contains two variable terms that govern the update behavior. Changing these two terms effectively defines
an infinite class of learning algorithms where  different values lead to significantly different results. We study this effect and propose a novel learning framework that improves over existing methods.

We evaluate our methods using the sequential question answering (SQA) dataset~\cite{Iyyer:17seq-qa}, and show that our proposed improvements to the search and update steps consistently enhance existing approaches.
The proposed algorithm achieves new state-of-the-art and outperforms existing parsers by 5.0\%.

  \section{Background}	%
\label{sec:background}

We give a formal problem definition of the semantic parsing task, followed by
the general learning framework for solving it.

\subsection{The Semantic Parsing Task}

The problem discussed in this paper can be formally defined as follows.
Let $\allinstruction$ be the set of all possible questions, $\programset$ programs (e.g., SQL-like queries), 
$\alltable$ tables (i.e., the structured data in this work) 
and $\allanswer$ answers.
We further assume access to an executor $\exec: \programset \times \alltable \rightarrow \allanswer$, that given a program $\program \in \programset$ and a table $\tab \in \alltable$, generates an answer $\exec(\program, \tab) \in \allanswer$.
We assume that the executor and all tables are deterministic and the executor can be called as many times as possible.
To facilitate discussion in the following sections, we define an environment function $\env_{\tab}: \programset \rightarrow \allanswer$, by applying the executor to the program as $\env_{\tab}(\program) = \exec(\program, \tab)$.

Given a question $\instruction$ and an environment $\env_{\tab}$, our aim is to \emph{generate} a program $\program^* \in \programset$ and then execute it to produce the answer $\env_{\tab}(\program^*)$. %
Assume that for any $\program \in \programset$, the score of $\program$ being a correct program for $\instruction$ is $\score_\theta(\program, \instruction, \tab)$, parameterized by $\theta$. The \emph{inference} task is thus:
 \begin{equation}
  \program^* = \argmax_{\program \in \programset} \score_\theta(\program, \instruction, \tab)
  \label{eq:search}
\end{equation}
As the size of $\programset$ is exponential to the length of the program, a generic \emph{search} procedure is typically
employed for Eq.~\eqref{eq:search}, as efficient dynamic algorithms typically do not exist.
These search procedures generally maintain a beam of program states sorted according to some scoring function, where each program state represents an incomplete program. The search then generates a new program state from an existing state by performing an action. Each action adds a set of tokens (e.g., {\tt Nation}) and keyword (e.g., {\tt Select}) to a program state. For example, in order to generate the program in Figure~\ref{fig:example}, the DynSP parser~\cite{Iyyer:17seq-qa} will take the first action as adding the SQL expression $\texttt{Select Nation}$.
Notice that $\score_\theta$ can be used in either probabilistic or non-probabilistic models. For probabilistic models, we assume that it is a Boltzmann policy, meaning that $p_\theta(\program \mid \instruction, \tab) \propto \exp\{\score_\theta(\program, \instruction, \tab)\}$. 

\subsection{Learning}

Learning a semantic parser is equivalent to learning the parameters $\theta$ in the scoring function, which is a 
structured learning problem, due to the large, structured output space $\programset$.
Structured learning algorithms generally consist of two major components: \emph{search} and \emph{update}. When the gold programs are available during training, the search procedure finds a set of high-scoring incorrect programs. These programs are used by the update step to derive loss for updating parameters. 
For example, these programs are used for approximating the partition-function in maximum-likelihood objective~\cite{Liang:11}
and finding set of programs causing margin violation in margin based methods~\cite{daume2005learning}.
Depending on the exact algorithm being used, these two components are not necessarily separated into isolated steps. For instance, parameters can be updated in the middle of search (e.g.,~\citealp{huang2012structured}).

For learning semantic parsers from denotations, where we assume only answers are available in a training set $\{(\instruction_i, \tab_i, \answer_i)\}_{i=1}^N$ of $N$ examples, the basic construction of the learning algorithms remains the same. However, the problems that search needs to handle
in SpFD is more challenging. In addition to finding a set of high-scoring incorrect programs, the search procedure also needs to \emph{guess} the correct program(s) evaluating to the gold answer $\answer_i$.
This problem is further complicated by the presence of spurious programs, which generate the correct answer 
but are semantically incompatible with the question. For example, although all programs in Figure~\ref{fig:flowchart}  evaluate to the same answer, only one of them is correct.
The issue of the spurious programs also affects the design of model update. For instance, maximum marginal likelihood methods treat all the programs that evaluate to the gold answer equally, while maximum margin reward networks use model score to break tie
and pick one of the programs as the correct reference.

\section{Addressing Spurious Programs: \mbox{ } Policy Shaping}

\label{sec:search}

Given a training example $(\instruction, \tab, \answer)$,  the aim of the search step is to find a set $\tt \mathcal{K}(\instruction, \tab, \answer)$ of programs consisting of correct programs that evaluate to $\answer$ and high-scoring incorrect programs. The search step should avoid picking up spurious programs for learning since such programs typically do not generalize. For example, in Figure~\ref{fig:flowchart}, the spurious program {\tt Select Nation Where Index is Min} will evaluate to an incorrect answer if the indices of the first two rows are swapped\footnote{This transformation preserves the answer of the question.}. 
This problem is challenging since among the programs that evaluate to the correct answer, most of them are spurious.

The search step can be viewed as following an exploration policy $b_\theta(\program | \instruction, \tab, \answer)$ to explore the set of programs $\programset$. 
This exploration is often performed by beam search and at each step, we either sample from $b_\theta$ or take the top scoring programs. The set $\tt \mathcal{K}(\instruction, \tab, \answer)$ is then used by the update step for parameter update. Most search strategies use an exploration policy which is based on the score function, for example %
$b_\theta(\program | \instruction, \tab, \answer) \propto \exp\{\score_\theta(\program, \tab)\}$. However, this approach can suffer from a \emph{divergence} phenomenon whereby the score of spurious programs picked up by the search in the first epoch increases, making it more likely for the search to pick them up in the future. Such \emph{divergence} issues are common with latent-variable learning and often require careful initialization to overcome~\cite{rose1998deterministic}. Unfortunately such initialization schemes are not applicable for deep neural networks which form the model of most successful semantic parsers today~\cite{Jia:16recombination,Misra:16neuralccg,Iyyer:17seq-qa}. Prior work, such as $\epsilon$-greedy exploration~\cite{Guu:17rl-mml}, has reduced the severity of this problem by introducing random noise in the search procedure to avoid saturating the search on high-scoring spurious programs.  However, random noise need not bias the search towards the correct program(s). In this paper, we introduce a simple policy-shaping method to guide the search. This approach allows incorporating prior knowledge in the exploration policy and can bias the search away from spurious programs.

\paragraph{Policy Shaping} Policy shaping is a method to introduce prior knowledge into a policy~\cite{Griffith:13policyshaping}. Formally, let the current behavior policy be $b_\theta(\program | \instruction, \tab, 
\answer)$ and a predefined critique policy, the prior knowledge, be $\policy_c(\program | \instruction, \tab)$. Policy shaping defines a new \emph{shaped behavior policy}  $\policy_b(\program | \instruction, \tab)$ given by:
\begin{equation}
\policy_b(\program | \instruction, \tab) = \frac{b_\theta(\program | \instruction, \tab, \answer) \policy_c(\program | \instruction, \tab)}{\sum_{\program' \in \programset}b_\theta(\program' | \instruction, \tab, \answer) \policy_c(\program' | \instruction, \tab)}.
\end{equation}

Using the shaped policy for exploration biases the search towards the critique policy's preference. We next describe a simple critique policy that we use in this paper.

\begin{figure}[t!]
\begin{tabular}{@{}p{7cm}}
\vspace{-0.275in}
\begin{algorithm}[H]
\caption{Learning a semantic parser from denotation using generalized updates.}
\begin{algorithmic}[1]
\footnotesize
\Require Training set $\{(\instruction_i, \tab_i, \answer_i\}_{i=1}^N$~(see Section~\ref{sec:background}), learning rate $\mu$ and stopping epoch $T$\~(see Section~\ref{sec:learning}).
\Definitions $\score_\theta(\program, \instruction, \tab)$ is a semantic parsing model parameterized by $\theta$. $p_s(\program \mid \instruction, \tab)$ is the policy used for exploration and $\text{search}(\theta, \instruction, \tab, \answer, p_s)$ generates candidate programs for updating parameters~(see Section~\ref{sec:search}). $\Delta$ is the generalized update~(see Section~\ref{sec:learning}). 
\Ensure Model parameters $\theta$.
\State \Comment{Iterate over the training data.}
\For{$t=1$ to $T$, $i=1$ to $N$}\label{alg:learn:dataloop:start}
\State \Comment{Find candidate programs using the shaped policy.}
\State $\mathcal{K} = \text{search}(\theta, \instruction_i, \tab_i, \answer_i, p_s)$\label{alg:learn:search}
\State \Comment{Compute generalized gradient updates}
\State $\theta = \theta + \mu \Delta( \mathcal{K})$\label{alg:learn:paramupdate}
\EndFor\label{alg:learn:dataloop:end}
\State \textbf{return} $\theta$ \label{alg:main:return}
\end{algorithmic}
\label{alg:learn} 
\end{algorithm}
\end{tabular}
\end{figure}

\paragraph{Lexical Policy Shaping} We qualitatively observed that correct programs often contains tokens which are also present in the question. For example, the correct program in Figure~\ref{fig:flowchart} contains the token \nlstring{Points}, which is also present in the question. We therefore, define a simple surface form similarity feature ${\tt match}(\instruction, \program)$ that computes the ratio of number of non-keyword tokens in the program $\program$ that are also present in the question $\instruction$. 

However, surface-form similarity is often not enough. For example, both the first and fourth program in Figure~\ref{fig:flowchart} contain the token \nlstring{Points} but only the fourth program is correct. Therefore, we also use a simple co-occurrence feature that triggers on frequently co-occurring pairs of tokens in the program and instruction. For example, the token \nlstring{most} is highly likely to co-occur with a correct program containing the keyword ${\tt Max}$. %
This happens for the example in Figure~\ref{fig:flowchart}. Similarly the token \nlstring{not} may co-occur with the keyword ${\tt Not\-Equal}$. %
We assume access to a lexicon $\Lambda = \{(\token_j, \predicate_j)\}_{j=1}^k$ containing $k$ lexical pairs of tokens and keywords. Each lexical pair $(\token, \predicate)$ maps the token $\token$ in a text to a keyword $\predicate$ in a program. For a given program $\program$ and question $\instruction$, we define a co-occurrence score as ${\tt co\_occur}(\program, \instruction) = \sum_{(\token, \predicate) \in \Lambda}\mathbbm{1}\{\token \in \instruction \land \predicate \in \program \}\}$. We define critique score ${\tt critique}(\program, \instruction)$ as the sum of the ${\tt match}$ and ${\tt co\_occur}$ scores. The critique policy is given by:
\begin{equation}
\policy_c(\program | \instruction, \tab) \propto \exp\left(\eta * {\tt critique}(\program, \instruction) \right),
\end{equation}
where $\eta$ is a single scalar hyper-parameter denoting the confidence in the critique policy. 

\section{Addressing Update Strategy Selection: Generalized Update Equation}	%

\label{sec:learning}

Given the set of programs generated by the search step, one can use many objectives to update the parameters. For example, previous work have utilized maximum marginal likelihood~\cite{krishnamurthy2017neural,Guu:17rl-mml}, reinforcement learning~\cite{zhongSeq2SQL2017,Guu:17rl-mml} and margin based methods~\cite{Iyyer:17seq-qa}. It could be difficult to choose the suitable algorithm from these options.

In this section, we propose a principle and general update equation
such that previous update algorithms can be considered
as special cases to this equation. Having a general update is important for the following reasons. First, it allows us to understand existing algorithms better by examining their basic properties. Second, the generalized update equation also makes it easy to implement and experiment with various different algorithms. Moreover, it provides a framework that enables the development of new variations or extensions of existing learning methods.

In the following, we 
describe how the commonly used algorithms are in fact very similar -- 
their update rules can all be viewed as special cases of the proposed generalized update equation.  Algorithm~\ref{alg:learn} shows the meta-learning framework. For every training example, we first find a set of candidates using an exploration policy~(line~\ref{alg:learn:search}). We use the program candidates to update the parameters~(line~\ref{alg:learn:paramupdate}).

\subsection{Commonly Used Learning Algorithms}
\label{sec:algm}

We briefly describe three algorithms: \emph{maximum marginalized likelihood}, \emph{policy gradient} and \emph{maximum margin reward}. 

\paragraph{Maximum Marginalized Likelihood} The maximum marginalized likelihood method maximizes the log-likelihood of the training data by marginalizing over the set of programs.
\begin{eqnarray}
J_{MML} &=& \log \policy(\answer_i | \instruction_i, \tab_i)  \nonumber \\
 &=& \log\sum_{\program \in \programset} \policy(\answer_i | \program, \tab_i)\policy(\program | \instruction_i, \tab_i)
\end{eqnarray}

Because an answer is deterministically computed given a program and a table, we define  $\policy(\answer \mid \program, \tab)$ as 1 or 0 depending upon whether the $\program$ evaluates to $\answer$ given $\tab$, or not.
Let ${\tt Gen(\answer, \tab)} \subseteq \programset$ be the set of compatible programs that evaluate to $\answer$ given the table $\tab$. The objective can then be expressed as:
\begin{eqnarray}
J_{MML} &=& \log \sum_{y \in {\tt Gen(\answer_i, \tab_i)}} \policy(\program | \instruction_i, \tab_i)
\end{eqnarray}
In practice, the summation over ${\tt Gen(.)}$ is approximated by only using the compatible programs in the set $\mathcal{K}$ generated by the search step.

\paragraph{Policy Gradient Methods} Most reinforcement learning approaches for semantic parsing assume access to a reward function $R: \programset \times \allinstruction \times \allanswer \rightarrow \mathbb{R},$ giving a scalar reward $R(\program, \answer)$ for a given program $\program$ and the correct answer $\answer$.\footnote{This is essentially a contextual bandit setting. \citet{Guu:17rl-mml} also used this setting. A general reinforcement learning setting requires taking a sequence of actions and receiving a reward for each action. For example, a program can be viewed as a sequence of parsing actions, where each action can get a reward. We do not consider the general setting here.} We can further assume without loss of generality that the reward is always in $[0,1]$. %
Reinforcement learning approaches maximize the expected reward $J_{RL}$:
\begin{eqnarray}
J_{RL} = \sum_{\program \in \programset} \policy(\program | \instruction_i, \tab_i) R(\program, \answer_i)
\label{eq:jrl}
\end{eqnarray}
$J_{RL}$ is hard to approximate using numerical integration since the reward for all programs may not be known a priori. Policy gradient methods solve this by approximating the derivative using a sample from the policy. When the search space is large, the policy may fail to sample a correct program, which can greatly slow down the learning. Therefore, off-policy methods are sometimes introduced to bias the sampling towards high-reward yielding programs. In those methods, an additional exploration policy $u(\program | \instruction_i, \tab_i, \answer_i)$ is used to improve sampling. Importance weights are used to make the gradient unbiased~(see Appendix for derivation). %

\paragraph{Maximum Margin Reward} For every training example $(\instruction_i,\tab_i, \answer_i)$, the maximum margin reward method finds the highest scoring program $\program_i$ that evaluates to $\answer_i$, as the \emph{reference} program, from the set $\mathcal{K}$ of programs generated by the search. With a margin function $\delta: \programset \times \programset \times \allanswer \rightarrow \mathbb{R}$ and reference program $\program$, 
the set of programs $\mathcal{V}$ that violate the margin constraint can thus be defined as:
\begin{eqnarray} 
\mathcal{V} = \{\program' \mid \program'\in \programset \,\,\textbf{and}\,\,\score_\theta(\program, \instruction, \tab) \nonumber \\
& \hspace{-2in} \le \score_\theta(\program', \instruction, \tab) + \delta(\program, \program', \answer)\} ,
\label{eq:violation}
\end{eqnarray}
where $\delta(\program, \program', \answer) = R(\program,\answer) -R(\program',\answer)$.
Similarly, the program that most violates the constraint can be written as:
\begin{eqnarray}
\bar{\program} = \arg\max_{\program' \in \programset}\{\score_\theta(\program', \instruction, \tab) + \delta(\program, \program', \answer) \nonumber \\
& \hspace{-2in} - \score_\theta(\program, \instruction, \tab)\} \label{eq:most-violate}
\end{eqnarray}
The most-violation margin objective (negative margin loss) is thus defined as:
\begin{eqnarray*}
J_{MMR} &=& -\max\{0, \score_\theta(\bar{\program}, \instruction_i, \tab_i) \\
&& - \score_\theta(\program_i, \instruction_i, \tab_i) + \delta(\program_i, \bar{\program}, \answer_i)\}
\end{eqnarray*}
Unlike the previous two learning algorithms, margin methods only update the score of the reference program and the program that violates the margin.

\subsection{Generalized Update Equation} 

\begin{table*}
\centering
\textbf{Generalized Update Equation:}
\begin{equation}
\label{eqn:unified}
\Delta(\mathcal{K}) = \sum_{\program \in \mathcal{K}} w(\program, \instruction, \tab, \answer)\left( \nabla_\theta\score_\theta(\program, \instruction, \tab) - \sum_{\program' \in \programset} q(\program' | \instruction, \tab) \nabla_\theta \score_\theta(\program', \instruction, \tab)\right) 
\end{equation} \\
\begin{tabular}{|l|c|c|}
\hline
\textbf{Learning Algorithm} & \textbf{Intensity}  & \textbf{Competing Distribution} \\
& $w(\program, \instruction, \tab, \answer)$ & $q(\program | \instruction, \tab)$ \\
\hline \hline
Maximum Margin Likelihood & $\frac{\policy(\answer | \program)\policy(\program | \instruction)}{\sum_{\program'} \policy(\answer | \program')\policy(\program' | \instruction )}$ & $\policy(\program | \instruction)$ \\
\hline
Meritocratic($\beta$) & $\frac{(\policy(\answer | \program)\policy(\program | \instruction))^\beta}{\sum_{\program'} (\policy(\answer | \program')\policy(\program' | \instruction ))^\beta}$ & $\policy(\program | \instruction)$ \\
\hline
REINFORCE & $\mathbbm{1}\{\program = \hat{\program}\} R(\program, \answer)$ & $\policy(\program | \instruction)$ \\
\hline
Off-Policy Policy Gradient & $\mathbbm{1}\{\program = \hat{\program}\}\, R(\program, \answer)\frac{\policy(\program | \instruction)}{u(\program | \instruction, \answer)}$ & $\policy(\program | \instruction)$ \\
\hline
Maximum Margin Reward (MMR) &  $\mathbbm{1}\{\program = \program^*\}$ & $\mathbbm{1}\{\program = \bar{\program}\}$ \\
\hline 
{\footnotesize Maximum Margin Avg.~Violation Reward} (MAVER) & $\mathbbm{1}\{\program = \program^*\}$ & $1/|\mathcal{V}|\mathbbm{1}\{\program \in \mathcal{V}\}$ \\
\hline
\end{tabular}
\caption{Parameter updates for various learning algorithms are special cases of Eq.~\eqref{eqn:unified}, with different choices of intensity $w$ and competing distribution $q$. We do not show dependence upon table $\tab$ for brevity. For off-policy policy gradient, $u$ is the exploration policy. For margin methods, $\program^*$ is the reference program (see \S\ref{sec:algm}), $\mathcal{V}$ is the set of programs that violate the margin constraint (cf.~Eq.~\eqref{eq:violation}) and $\bar{\program}$ is the most violating program (cf.~Eq.~\eqref{eq:most-violate}). For REINFORCE, $\hat{\program}$ is sampled from $\mathcal{K}$ using $p(.)$ whereas for Off-Policy Policy Gradient, $\hat{\program}$ is sampled using $u(.)$.}
\label{tbl:unified}
\end{table*}

Although the algorithms described in \S\ref{sec:algm} seem very different on the surface, the gradients of their loss
functions can in fact be described in the same generalized form, given in Eq.~\eqref{eqn:unified}\footnote{See Appendix for the detailed derivation.}. In addition to the gradient of the model scoring function, this equation has two variable terms, $w(\cdot)$, $q(\cdot)$. 
We call the first term $w(\program, \instruction,\tab, \answer)$ \emph{intensity}, which is a positive scalar value and the second term $q(\program | \instruction, \tab)$ the \emph{competing distribution}, which is a probability distribution over programs. 
Varying them makes the equation equivalent to the update rule of the algorithms we discussed, as shown in Table~\ref{tbl:unified}. We also consider meritocratic update policy which uses a hyperparameter $\beta$ to sharpen or smooth  the intensity of maximum marginal likelihood~\cite{Guu:17rl-mml}. 

Intuitively, $w(\program, \instruction, \tab, \answer)$ defines the positive part of the update equation, which defines how aggressively the update favors program $\program$. Likewise, $q(\program | \instruction, \tab)$ defines the negative part of the learning algorithm, namely how aggressively the update penalizes the members of the program set.

The generalized update equation 
provides a tool for better understanding  individual algorithm, and helps shed some light on when a particular method may perform better. 

\paragraph{Intensity versus Search Quality}
In SpFD, the effectiveness of the algorithms for SpFD is closely related to the quality of the search results given that the gold program is not available. Intuitively, if
the search quality is good, the update algorithm could be aggressive on updating the model parameters. When the search quality is poor, the algorithm should be conservative.

The intensity $w(\cdot)$ is closely related to the aggressiveness of the algorithm.
For example, the maximum marginal likelihood is less aggressive given that
it produces a non-zero intensity over all programs in the program set $\mathcal{K}$ that evaluate to the correct answer. The intensity for a particular correct program $\program$ is proportional to its probability $p(\program | \instruction, \tab)$. Further, meritocratic update becomes more aggressive as $\beta$ becomes larger.

In contrast, REINFORCE and maximum margin reward both have a non-zero intensity only on a single program in $\mathcal{K}$. This value is 1.0 for maximum margin reward, while for reinforcement learning, this value is the reward. Maximum margin reward therefore updates most aggressively in favor of its selection while maximum marginal likelihood tends to hedge its bet. Therefore, the maximum margin methods should benefit the most when the search quality improves. %

\paragraph{Stability}
The general equation also allows us to investigate the stability of a model update algorithm.
In general, the variance of update direction can be high, hence less stable, if the
model update algorithm has peaky competing distribution, or it puts all of its intensity on a single program. 
For example, REINFORCE only samples one
program and puts non-zero intensity only on that program, so it could be unstable depending
on the sampling results.

The competing distribution affects the stability of the algorithm. For example, maximum margin reward penalizes only the most violating program and is benign to other incorrect programs. Therefore, the MMR algorithm could be unstable during training.

\paragraph{New Model Update Algorithm}
The general equation provides a framework that enables the development of new variations or extensions of existing learning methods.
For example, in order to improve the stability of the MMR algorithm, we 
propose a simple variant of maximum margin reward, which penalizes all violating programs instead of only the most violating one. We call this approach \emph{maximum margin average violation reward} (MAVER), which is included in Table~\ref{tbl:unified} as well.
Given that MAVER effectively considers more negative examples during each update, we expect that it is more stable compared to the MMR algorithm.

  \section{Experiments} %
\label{sec:exp}

We describe the setup in \S\ref{sec:setup} and results in \S\ref{sec:results}.

\subsection{Setup}
\label{sec:setup}

\paragraph{Dataset}

We use the sequential question answering (SQA) dataset~\cite{Iyyer:17seq-qa} for our experiments. SQA contains 6,066 sequences and each sequence contains up to 3 questions, with 17,553 questions in total. The data is partitioned into training (83\%) and test (17\%) splits. We use 4/5 of the original train split as our training set and the remaining 1/5 as the dev set. We evaluate using exact match on answer.
Previous state-of-the-art result on the SQA dataset is 44.7\% accuracy, using maximum margin reward learning.

\paragraph{Semantic Parser}

Our semantic parser is based on DynSP~\cite{Iyyer:17seq-qa}, which contains a set of SQL actions, such as adding a clause (e.g.,  {\tt Select Column}) or adding an operator (e.g., {\tt Max}). Each action has an associated neural network module that generates the score for the action based on the instruction, the table and the list of past actions. The score of the entire program is given by the sum of scores of all actions.

We modified DynSP to improve its representational capacity. We refer to the new parser as DynSP++. Most notably, we included new features and introduced two additional parser actions. See  Appendix 8.2 for more details. While these improvements help us achieve state-of-the-art results, the majority of the gain comes from the learning contributions described in this paper.

\paragraph{Hyperparameters}

For each experiment, we train the model for 30 epochs. We find the optimal stopping epoch by evaluating the model on the dev set. We then train on train+dev set till the stopping epoch and evaluate the model on the held-out test set. Model parameters are trained using stochastic gradient descent with learning rate of $0.1$. We set the hyperparameter $\eta$ for policy shaping to 5. All hyperparameters were tuned on the dev set. We use 40 lexical pairs for defining the {\tt co-occur} score. We used common English superlatives (e.g., \emph{highest}, \emph{most}) and comparators (e.g., \emph{more}, \emph{larger}) and did not fit the lexical pairs based on the dataset. 

Given the model parameter $\theta$, we use a base exploration policy defined in~\cite{Iyyer:17seq-qa}. This exploration policy is given by $b_\theta(\program \mid \instruction, \tab, \answer) \propto \exp(\lambda \cdot R(\program, \answer) + \score_\theta(\program, \theta, \answer))$. $R(\program, \answer)$ is the reward function of the incomplete program $\program$, given the answer $\answer$. We use a reward function $R(y, z)$ given by the Jaccard similarity of the gold answer $z$ and the answer generated by the program $y$. The value of $\lambda$ is set to infinity, which essentially is equivalent to sorting the programs based on the reward and using the current model score for tie breaking. Further, we prune all syntactically invalid programs. For more details, we refer the reader to~\cite{Iyyer:17seq-qa}.

\begin{table*}
\centering
\begin{tabular}{|c|c|c|c|c|}
  \hline
  \textbf{Algorithm} & \multicolumn{2}{c|}{\textbf{Dev}} & \multicolumn{2}{c|}{\textbf{Test}} \\
  \cline{2-5}
  {}            & w.o. Shaping & w. Shaping & w.o. Shaping & w. Shaping \\
  \hline
  Maximum Margin Likelihood           &    33.2      &  32.5      &   31.0      &   32.3    \\
  \hline
  Meritocratic $(\beta=0)$ & 27.1 & 28.1 & 31.3 & 30.1 \\
  \hline
  Meritocratic $(\beta=0.5)$ & 28.3 & 28.7 & 31.7 & 32.0 \\
  \hline
  Meritocratic $(\beta=\infty)$ & 39.3 & 41.6 & 41.6& 45.2 \\
  \hline 
  \hline
  REINFORCE            & 10.2         & 11.8       & 2.4          & 4.0        \\
\hline
Off-Policy Policy Gradient & 36.6         & 38.6       & 42.6         & 44.1       \\
\hline
\hline
MMR           & 38.4         & 40.7       & 43.2         & 46.9       \\
\hline
MAVER         &    39.6      &   44.1     & 43.7         & {\bf 49.7}       \\

\hline
\end{tabular}
\caption{Experimental results on different model update algorithms, with and without policy shaping.}
\label{tbl:dynsp-results}
\end{table*}

\subsection{Results}
\label{sec:results}

Table~\ref{tbl:dynsp-results} contains the dev and test results when using our algorithm on the SQA dataset.
We observe that margin based methods perform better than maximum likelihood methods and policy gradient in our experiment. Policy shaping in general improves
the performance across different algorithms. Our best test results outperform
previous SOTA by 5.0\%.

\paragraph{Policy Gradient vs Off-Policy Gradient}
REINFORCE, a simple policy gradient method, achieved extremely poor performance. This likely due to the problem of exploration and having to sample from a large space of programs. This is further corroborated from observing the much superior performance of off-policy policy gradient methods. Thus, the sampling policy is an important factor
to consider for policy gradient methods.

\paragraph{The Effect of Policy Shaping}
We observe that the improvement due to policy shaping is 6.0\% on the SQA dataset for MAVER and only 1.3\% for maximum marginal likelihood. We also observe that as $\beta$ increases, the improvement due to policy shaping for meritocratic update increases. This supports our hypothesis that aggressive updates of margin based methods is beneficial when the search method is more accurate as compared to maximum marginal likelihood which hedges its bet between all programs that evaluate to the right answer.

\paragraph{Stability of MMR}
In Section~\ref{sec:learning}, the general update equation helps us point out that MMR could be unstable due
to the peaky competing distribution. MAVER was proposed
to increase the stability of the algorithm. To measure stability, we calculate
the mean absolute difference of the development set accuracy between successive epochs during
training, as it indicates how much an algorithm's performance fluctuates during training.
With this metric, we found mean difference for MAVER is 0.57\% where the
mean difference for MMR is 0.9\%. This indicates that MAVER is in fact
more stable than MMR.

\paragraph{Other variations}
We also analyze other possible novel learning algorithms that are made possible due to generalized update equations. Table~\ref{tbl:novel-combination} reports development results using these algorithms.  By mixing different intensity scalars and competing distribution
from different algorithms, we can create new variations of the model update algorithm. In Table~\ref{tbl:novel-combination}, we show that
by mixing the MMR's intensity and MML's competing distribution,
we can create an algorithm that outperform MMR on  the development set.

\begin{table}[t]
\centering
\begin{tabular}{|l|l|c|c|c|}
\hline
\textbf{$w$} & \textbf{$q$} & Dev\\
\hline
MMR & MML &  {\bf 41.9} \\
Off-Policy Policy Gradient & MMR &  37.0 \\ \hline
MMR & MMR & 40.7 \\
\hline
\end{tabular}
\caption{The dev set results on the new variations of the update algorithms.
}
\label{tbl:novel-combination}
\end{table}

\paragraph{Policy Shaping helps against Spurious Programs} In order to better understand if policy shaping helps bias the search away from spurious programs, we analyze 100 training examples. We look at the highest scoring program in the beam at the end of training using MAVER. Without policy shaping, we found that 53 programs were spurious while using policy shaping this number came down to 23. We list few examples of spurious program errors corrected by policy shaping in Table~\ref{label:spurious-corrections}.
\begin{table*}
\centering
\begin{tabular}{|c|c|c|}
\hline
\textbf{Question} & \textbf{without policy shaping} & \textbf{with policy shaping}\\
\hline
\emph{``of these teams, which had more } & {{ {\tt SELECT} {\tt Club}}} & {{ {\tt SELECT} {\tt Club}}}\\
\emph{than 21 losses?"}& { {\tt WHERE} {\tt Losses} {\tt=} {\tt ROW 15}} & { {\tt WHERE} {\tt Losses} {\tt >} {\tt 21}} \\
\hline
\emph{``of the remaining, which } & {\tt SELECT Nation WHERE} & {\tt FollowUp WHERE} \\
\emph{earned the most bronze medals?"} & {\tt Rank = ROW 1} & {\tt  Bronze is Max}\\
\hline
\emph{``of those competitors from germany, } & {\tt SELECT Name WHERE }& {\tt FollowUp WHERE } \\
\emph{which was not paul sievert?"} & {\tt Time (hand) = ROW 3}& {\tt Name != ROW 5}\\
\hline
\end{tabular}
\caption{Training examples and the highest ranked program in the beam search, scored according to the shaped policy, after training with MAVER. Using policy shaping, we can recover from failures due to spurious programs in the search step for these examples.}
\label{label:spurious-corrections}
\end{table*}

\paragraph{Policy Shaping vs Model Shaping} Critique policy contains useful information that can bias the search away from spurious programs. Therefore, one can also consider making the critique policy as part of the model. We call this model shaping. We define our model to be the shaped policy and train and test using the new model. Using MAVER updates, we found that the dev accuracy dropped to 37.1\%. We conjecture that the strong prior in the critique policy can hinder generalization in model shaping.

  \section{Related Work}%
\label{sec:related}

\paragraph{Semantic Parsing from Denotation} Mapping natural language text to formal meaning representation was first studied by~\citet{montague1970english}.
Early work on learning semantic parsers rely on labeled formal representations as the supervision signals~\cite{Zettlemoyer:05,Zettlemoyer:07,zelle1993learning}.
However, because getting access to gold formal representation generally requires expensive annotations by an expert, distant supervision approaches, where semantic parsers are learned from denotation only, have become the main learning paradigm~(e.g., \citealp{Clarke:10, Liang:11, Artzi:13,Berant:13,Iyyer:17seq-qa,krishnamurthy2017neural}).
\citet{Guu:17rl-mml} studied the problem of spurious programs and considered adding noise to diversify the search procedure and introduced meritocratic updates.

\paragraph{Reinforcement Learning Algorithms} Reinforcement learning algorithms have been applied to various NLP problems including dialogue~\cite{Li:16drlchatbot}, text-based games~\cite{Narasimhan:15text-games}, information extraction~\cite{Narasimhan:16rl-ie}, coreference resolution~\cite{Clark:16drl-coref}, semantic parsing~\cite{Guu:17rl-mml} and instruction following~\cite{Misra:17instructions}. \citet{Guu:17rl-mml} show that policy gradient methods underperform maximum marginal likelihood approaches. Our result on the SQA dataset supports their observation. However, we show that using off-policy sampling, policy gradient methods can provide superior performance to  maximum marginal likelihood methods.

\paragraph{Margin-based Learning} Margin-based methods have been considered in the context of SVM learning. In the NLP literature, margin based learning has been applied to parsing ~\cite{taskar2004max,McDonald2005OnlineLT}, text classification~\cite{Taskar2003MaxMarginMN}, machine translation~\cite{Watanabe2007OnlineLT} and semantic parsing~\cite{Iyyer:17seq-qa}. \citet{Kummerfeld2015AnEA} found that max-margin based methods generally outperform likelihood maximization on a range of tasks. Previous work have studied connections between margin based method and likelihood maximization for supervised learning setting. We show them as special cases of our unified update equation for  distant supervision learning. 
Similar to this work, \citet{kentonccg16} also found that in the context of supervised learning, margin-based algorithms which update all violated examples perform better than the one that only updates the most violated example.

\paragraph{Latent Variable Modeling} Learning semantic parsers from denotation can be viewed as a latent variable modeling problem, where the program is the latent variable. Probabilistic latent variable models have been studied using EM-algorithm and its variant~\cite{dempster1977maximum}.  The graphical model literature has studied latent variable learning on margin-based methods~\cite{yu2009learning} and probabilistic models~\cite{quattoni2007hidden}. \citet{samdani2012unified} studied various variants of EM algorithm and showed that all of them are special cases of a unified framework. Our generalized update framework is similar in spirit.  

  \section{Conclusion} %
\label{sec:conclusion}
In this paper, we propose a general update equation from semantic
parsing from denotation and propose a policy shaping method for addressing
the spurious program challenge. For the future, we plan to apply the proposed learning framework to more semantic parsing tasks and consider new methods for policy shaping.
  
  \section{Acknowledgements}

We thank Ryan Benmalek, Alane Suhr, Yoav Artzi, Claire Cardie, Chris Quirk, Michel Galley and members of the Cornell NLP group for their valuable comments. We are also grateful to Allen Institute for Artificial Intelligence for the computing resource support. This work was initially started when the first author interned at Microsoft Research.
  
\bibliographystyle{acl_natbib_nourl}
 \bibliography{main}
 \onecolumn
\section*{Appendix}

\subsection{Deriving the updates of common algorithms}

Below we derive the gradient of various learning algorithms. We assume access to a training data $\{(\instruction_i, \tab_i, \answer_i)\}_{i=1}^N$ with $N$ examples. Given an input instruction $\instruction$ and table $\tab$, we model the score of a program using a score function $\score_\theta(\program, \instruction, \answer)$ with parameters $\theta$. When the model is probabilistic, we assume it is a Boltzmann distribution given by $\policy(\program \mid \instruction, \tab) \propto \exp\{\score_\theta(\program, \instruction, \tab)\}$.

In our result, we will be using.
\begin{equation}
\label{eq:log_grad}
\nabla_\theta \log \policy(\program \mid \instruction, \tab) = \nabla_\theta \score_\theta(\program, \instruction, \tab) - \sum_{\program' \in \programset } \policy(\program' \mid \instruction, \tab) \nabla_\theta \score_\theta(\program', \instruction, \tab)
\end{equation}

\paragraph{Maximum Marginal Likelihood} The maximum marginal objective $J_{MML}$ can be expressed as:

\begin{equation*}
J_{MML} = \sum_{i=1}^N \log \sum_{\program \in {\tt Gen}(\tab_i, \answer_i)}\policy(\program \mid \instruction_i, \tab_i)
\end{equation*}

where ${\tt Gen}(\tab, \answer)$ is the set of all programs from $\programset$ that generate the answer $\answer$ on table $\tab$. Taking the derivative gives us:

\begin{eqnarray*}
\nabla_\theta J_{MML} &=& \sum_{i=1}^N \nabla_\theta \log \sum_{\program \in {\tt Gen}(\tab_i, \answer_i)} \policy(\program \mid \instruction_i, \tab_i)\\
 &=&\sum_{i=1}^N \frac{\sum_{\program \in {\tt Gen}(\tab_i, \answer_i)} \nabla_\theta  \policy(\program \mid \instruction_i, \tab_i)}{\sum_{\program \in {\tt Gen}(\tab_i, \answer_i)} \policy(\program \mid \instruction_i, \tab_i)}\\
\end{eqnarray*}

Then using Equation~\ref{eq:log_grad}, we get:
\begin{equation}
\nabla_\theta J_{MML} = \sum_{i=1}^N \sum_{\program \in {\tt Gen}(\tab_i, \answer_i)} w(\program \mid \instruction_i, \tab_i)  \left\{\nabla_\theta \score_\theta(\program, \instruction, \tab) - \sum_{\program' \in \programset } \policy(\program' \mid \instruction, \tab) \nabla_\theta \score_\theta(\program', \instruction, \tab)\right\}\\
\end{equation}

where $$w(\program, \instruction, \tab) = \frac{\policy(\program \mid \instruction, \tab)}{\sum_{\program' \in {\tt Gen}(\tab, \answer)}\policy(\program' \mid \instruction, \tab)}$$

\paragraph{Policy Gradient Methods}

Reinforcement learning based approaches maximize the expected reward objective.

\begin{equation}
J_{RL} = \sum_{i=1}^{N} \sum_{\program \in \programset} \policy(\program \mid \instruction_i, \tab_i) R(\program,\answer_i)
\end{equation}

We can then compute the derivate of this objective as:

\begin{equation}
\nabla_\theta J_{RL} = \sum_{i=1}^{N} \sum_{\program \in \programset} \nabla_\theta \policy(\program \mid \instruction_i, \tab_i) R(\program, \answer_i)
\end{equation}

The above summation can be expressed as expectation~\cite{Williams:92reinforce}.

\begin{equation}
\nabla_\theta J_{RL} = \sum_{i=1}^{N} \sum_{\program \in \programset} \policy(\program \mid \instruction_i, \tab_i) \nabla_\theta \log \policy(\program \mid \instruction_i, \tab_i) R(\program, \answer_i)
\end{equation}

For every example $i$, we sample a program $\program_i$ from $\programset$ using the policy $\policy(. \mid \instruction_i, \tab_i)$. In practice this sampling is done over the output programs of the search step. 

\begin{eqnarray*}
\nabla_\theta J_{RL} &\approx& \sum_{i=1}^{N} \nabla_\theta \log \policy(\program_i \mid \instruction_i, \tab_i) R(\program_i, \answer_i) \\
&& \mbox{using gradient of } \log \policy(.\mid .) \\
&\approx& \sum_{i=1}^{N} R(\program_i, \answer_i) \left\{\nabla_\theta \score_\theta(\program_i, \instruction_i, \tab) - \sum_{\program' \in \programset } \policy(\program' \mid \instruction_i, \tab_i) \nabla_\theta \score_\theta(\program', \instruction_i, \tab_i) \right\} \\
\end{eqnarray*}

\paragraph{Off-Policy Policy Gradient Methods}

In off-policy policy gradient method, instead of sampling a program using the current policy $\policy(.\mid .)$, we use a separate exploration policy $u(.\mid .)$. For the $i^{th}$ training example, we sample a program $\program_i$ from the exploration policy $u(.\mid \instruction_i, \tab_i, \answer_i)$. Thus the gradient of expected reward objective from previous paragraph can be expressed as:

\begin{eqnarray*}
\nabla_\theta J_{RL} &=& \sum_{i=1}^{N} \sum_{\program \in \programset} \nabla_\theta \policy(\program \mid \instruction_i, \tab_i) R(\program, \answer_i)\\
&=& \sum_{i=1}^{N} \sum_{\program \in \programset} u(\program \mid \instruction_i, \tab_i, \answer_i)\frac{\policy(\program \mid \instruction_i, \tab_i)}{u(\program \mid \instruction_i, \tab_i, \answer_i)} \nabla_\theta \log \policy(\program \mid \instruction_i, \tab_i) R(\program, \answer_i)\\
&& \mbox{using, for every } i\,\,\program_i \sim u(.\mid \instruction_i, \tab_i, \answer_i)\\
&\approx& \sum_{i=1}^{N}  \frac{\policy(\program \mid \instruction_i, \tab_i)}{u(\program \mid \instruction_i, \tab_i,\answer_i)} \nabla_\theta \log \policy(\program \mid \instruction_i, \tab_i) R(\program,  \answer_i)
\end{eqnarray*}

the ratio of $\frac{\policy(\program \mid \instruction, \tab)}{u(\program \mid \instruction, \tab,\answer)}$ is the importance weight correction. In practice, we sample a program from the output of the search step.

\paragraph{Maximum Margin Reward (MMR)}
For the $i^{th}$ training example, let $\mathcal{K}(\instruction_i, \tab_i, \answer_i)$ be the set of programs produced by the search step. Then MMR finds the highest scoring program in this set, which evaluates to the correct answer. Let this program be $y_i$. MMR optimizes the parameter to satisfy the following constraint:

\begin{equation}
\score_\theta(\program_i, \instruction_i, \tab_i) \ge \score_\theta(\program', \instruction_i, \tab_i) + \delta(\program_i, \program', \answer_i)\,\, \program' \in \programset
\end{equation}

where the margin $\delta(\program_i, \program', \answer_i)$ is given by $ R(\program_i, \answer_i) - R(\program', \answer_i)$. Let $\mathcal{V}$ be the set of violations given by: $\mathcal{V} = \{\score_\theta(\program', \instruction_i, \tab_i)  - \score_\theta(\program_i, \instruction_i, \tab_i) + \delta(\program_i, \program', \answer_i) > 0 \mid \program \in \programset\}$. 

At each training step, MMR only considers the program which is most violating the constraint. When $|\mathcal{V}| > 0$ then let $y^*$ be the most violating program given by: 

\begin{eqnarray*}
\bar{y} &=& \arg\max_{\program' \in \programset} \left\{\score_\theta(\program', \instruction_i, \tab_i)  - \score_\theta(\program_i, \instruction_i, \tab_i) + R(\program_i, \answer_i) - R(\program', \answer_i)\right\}\\
 &=& \arg\max_{\program' \in \programset} \left\{\score_\theta(\program', \instruction_i, \tab_i)  -  R(\program', \answer_i)\right\}
\end{eqnarray*}

Using the most violation approximation, the objective for MMR can be expressed as negative of hinge loss:

\begin{equation}
J_{MMR} = -\max\{0, \score_\theta(\bar{\program}, \instruction_i, \tab_i)  - \score_\theta(\program_i, \instruction_i, \tab_i) + R(\program_i, \answer_i) - R(\bar{\program}, \answer_i)\}
\end{equation}

Our definition of $\program^*$ allows us to write the above objective as:

\begin{equation}
J_{MMR} = -\mathbbm{1}\{\mathcal{V} > 0\}\{\score_\theta(\bar{\program}, \instruction_i, \tab_i)  - \score_\theta(\program_i, \instruction_i, \tab_i) + R(\program_i, \answer_i) - R(\bar{\program}, \answer_i)\}
\end{equation}

the gradient is then given by:

\begin{eqnarray}
\nabla_\theta J_{MMR} = -\mathbbm{1}\{\mathcal{V} > 0\}\{\nabla_\theta\score_\theta(\bar{\program}, \instruction_i, \tab_i) - \nabla_\theta \score_\theta(\program_i, \instruction_i, \tab_i) \}
\end{eqnarray}

\paragraph{Maximum Margin Average Violation Reward (MAVER)}

Given a training example, MAVER considers the same constraints and margin as MMR. However instead of considering only the most violated program, it considers
all violations. Formally, for every example $(\instruction_i, \tab_i, \answer_i)$ we compute the ideal program $\program_i$ as in MMR. We then optimize the average negative hinge loss error over all violations:

\begin{equation}
J_{MAVER} = -\frac{1}{
\mathcal{V}}\sum_{\program' \in \mathcal{V}}\{\score_\theta(\program', \instruction_i, \tab_i)  - \score_\theta(\program_i, \instruction_i, \tab_i) + R(\program_i, \answer_i) - R(\program', \answer_i)\}
\end{equation}

Taking the derivative we get:

\begin{eqnarray*}
\nabla_\theta J_{MAVER} &=& - \frac{1}{
\mathcal{V}}\sum_{\program' \in \mathcal{V}}\{\nabla_\theta\score_\theta(\program', \instruction_i, \tab_i)  - \nabla_\theta\score_\theta(\program_i, \instruction_i, \tab_i))\}\\
&=& \nabla_\theta\score_\theta(\program_i, \instruction_i, \tab_i)) - \sum_{\program' \in \mathcal{V}}\frac{1}{|\mathcal{V}|}\nabla_\theta\score_\theta(\program', \instruction_i, \tab_i)
\end{eqnarray*}

\subsection{Changes to DynSP Parser}
We make following 3 changes to the DynSP parser to increase its representational power. The new parser is called DynSP++. We describe these three changes below:

\begin{enumerate}
\item We add two new actions: disjunction ({\tt OR}) and follow-up cell ({\tt FpCell}). The disjunction operation is used to describe multiple conditions together example:

Question: \nlstring{what is the population of USA or China?}\\
Program: {\tt Select Population Where Name = China OR Name = USA}

Follow-up cell is only used for a question which is following another question and whose answer is a single cell in the table. Follow-up cell is used to select values for another column corresponding to this cell.

Question: \nlstring{and who scored that point?}\\
Program: {\tt Select Name Follow-Up Cell}

\item We add surface form features in the model for column and cell. These features trigger on token match between an entity in the table (column name or cell value) and a question. We consider two tokens: exact match and overlap. The exact match is 1.0 when every token in the entity is present in the question and 0 otherwise. Overlap feature is 1.0 when atleast one token in the entity is present in the question and 0 otherwise. We also consider related-column features that were considered by~\citet{krishnamurthy2017neural}.

\item We also add recall features which measure how many tokens in the question that are also present in the table are covered by a given program. To compute this feature, we first compute the set $\mathcal{E}_1$ of all tokens in the question that are also present in the table. We then find a set of non-keyword tokens $\mathcal{E}_2$ that are present in the program. The recall score is then given by $w *\frac{|\mathcal{E}_1 - \mathcal{E}_2|}{|\mathcal{E}_1|}$, where $w$ is a learned parameter.

\end{enumerate}

\end{document}